\documentclass{article}
\pdfoutput=1

\usepackage[preprint,nonatbib]{neurips_2023}

\usepackage[utf8]{inputenc} 
\usepackage[T1]{fontenc}    
\usepackage{hyperref}       
\usepackage{url}            
\usepackage{booktabs}       
\usepackage{amsfonts}       
\usepackage{nicefrac}       
\usepackage{microtype}      
\usepackage{xcolor}         
\usepackage{graphicx}
\usepackage{comment}
\usepackage{subcaption}
\usepackage{amssymb}
\usepackage{amsmath}

\title{MOD-CL: Multi-label Object Detection with Constrained Losss}

\author{%
  Sota Moriyama \\
  National Institute of Informatics\\
  Tokyo, Japan\\
  \texttt{sotam@nii.ac.jp} \\
  \And
  Koji Watanabe \\
  National Institute of Informatics\\
  Tokyo, Japan\\
  \texttt{kojiwatanabe@nii.ac.jp} \\\
  \And
  Katsumi Inoue \\
  National Institute of Informatics\\
  Tokyo, Japan\\
  \texttt{inoue@nii.ac.jp} \\\
  \And
  Akihiro Takemura \\
  National Institute of Informatics\\
  Tokyo, Japan\\
  \texttt{atakemura@nii.ac.jp} \\\
}

\begin{document}

\maketitle

\begin{abstract}
We introduce MOD-CL, a multi-label object detection framework that utilizes constrained loss in the training process to produce outputs that better satisfy the given requirements. In this paper, we use MOD\textsubscript{YOLO}, a multi-label object detection model built upon the state-of-the-art object detection model YOLOv8, which has been published in recent years. In Task 1, we introduce the Corrector Model and Blender Model, two new models that follow after the object detection process, aiming to generate a more constrained output. For Task 2, constrained losses have been incorporated into the MOD\textsubscript{YOLO} architecture using Product T-Norm. The results show that these implementations are instrumental to improving the scores for both Task 1 and Task 2.
\end{abstract}

\section{Introduction}
Object detection is a critical computer vision task that aims to identify the precise locations of objects in images or videos \cite{DBLP:journals/pieee/ZouCSGY23}. Due to the nature of this task, object detection has received a lot of attention in the context of autonomous driving. However, with autonomous driving, there is a need to have further information about the action the object is taking (action detection). On top of this, there has been some extensive research about making the outputs satisfy some given requirements \cite{DBLP:journals/ml/GiunchigliaSKCL23}. These are all crucial to the development of autonomous driving.

In this paper, we introduce MOD-CL, a multi-label object detection framework that utilizes constrained loss in the training process to better satisfy the requirements of the actions. Specifically, we develop a model called MOD\textsubscript{YOLO} that is based on YOLOv8 \cite{ultralytics}, a state-of-the-art object detection model released by Ultralytics. Additionally, we introduce two distinct models and training procedures to accommodate two different scenarios: (1) limited labeled data, and (2) the output of labels that satisfy the requirements. These will be referred to as Task 1 and Task 2 from here.

\section{YOLOv8 for Multi-labeled Object Detection}
As the original YOLOv8 only supported single labels per bounding box, we have modified the program to support multiple labels. To accomplish this, we used one-hot vector encodings, with multiple classes allowed to be set as true. Numerous parts of the original YOLOv8 model, including the input sequence and output sequences, have been modified. However, we will only touch on parts that directly affect the performance of the overall model, namely the Non-Maximum Suppression (NMS) algorithm that is used before outputting the predictions.

In the original YOLOv8 model, the outputs are produced through the use of NMS with regards to the confidence scores and IOU of each label for each bounding box. On the other hand, in our approach, we focused on outputting the bounding boxes with respect to the confidence scores of the agent labels. Specifically, the modifications done are as follows:
\begin{itemize}
    \item \textbf{Bounding Box Thresholding}: Only bounding boxes that have confidence scores for agent labels above the threshold are used.
    \item \textbf{Agent-wise NMS}: Excessive bounding boxes were reduced with the use of the NMS algorithm with respect to the bounding boxes and confidence scores of the agent labels.
\end{itemize}
This not only allows us to output multiple labels for every bounding box, but it also allows us to satisfy one of the pre-defined requirements: to have at least one agent label included.

As each of the bounding boxes has one label that corresponds to the type of agent, we focus on filtering with respect to the agent labels. This prevents the model from violating the requirement for the agents—to have at least one agent label included in the output. On top of all of the modifications, we also used an object tracking algorithm, BoT-SORT \cite{aharon2022bot} (already implemented in YOLOv8), to enhance our output accuracy.

We will refer to this variation of the YOLOv8 model with the above modifications, MOD\textsubscript{YOLO} (Multi-labeled Object Detection based on YOLO).

\section{Task 1}
\begin{figure}[t]
    \centering
    \begin{subfigure}{0.48\textwidth}
        \centering
    	\includegraphics[width=0.4\linewidth]{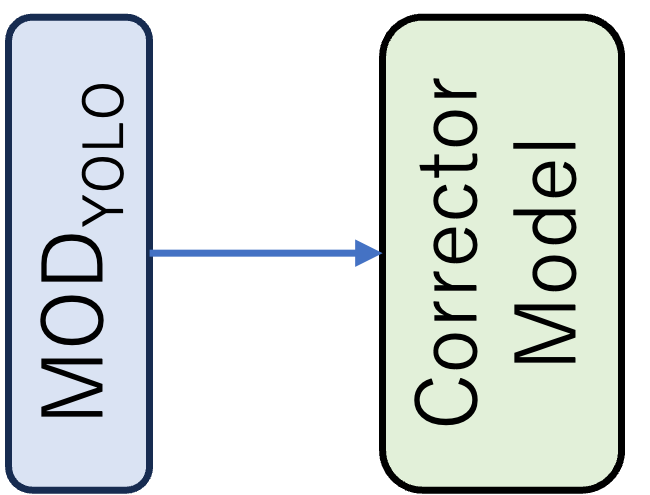}
    	\caption{Stage 1 of training}
    \end{subfigure}
    \begin{subfigure}{0.48\textwidth}
        \centering
    	\includegraphics[width=0.6\linewidth]{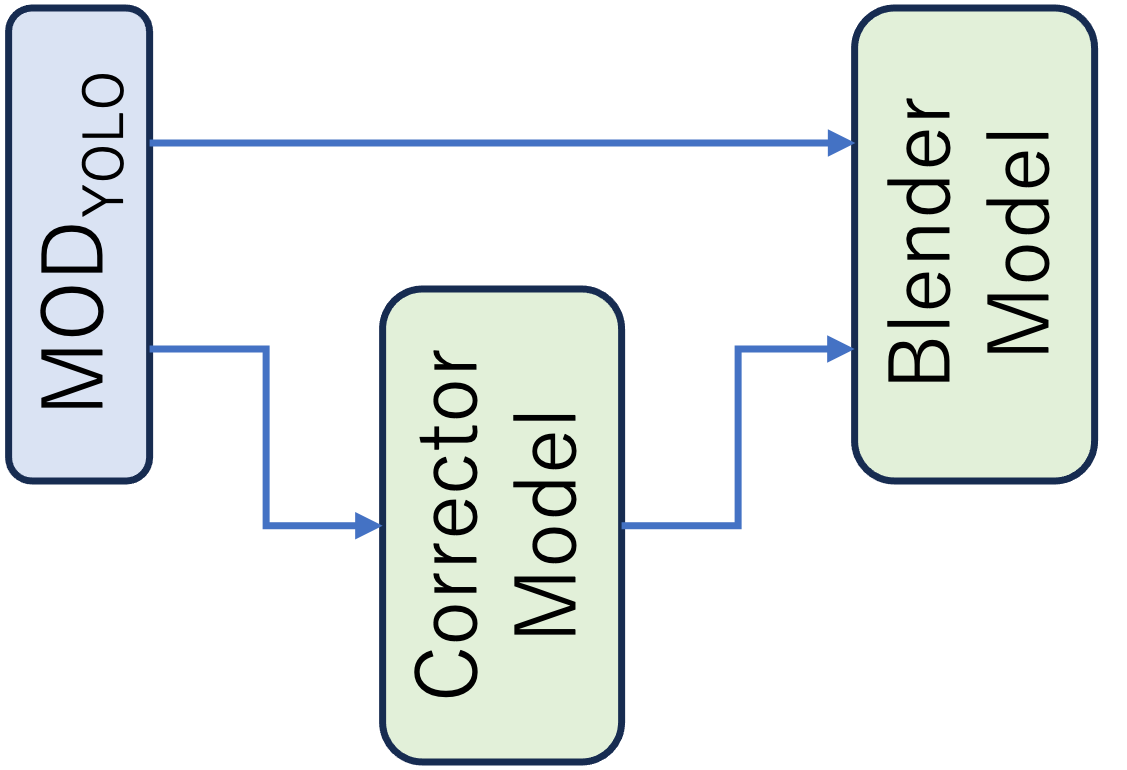}
    	\caption{Stage 2 of training}
    \end{subfigure}
    \caption{MOD\textsubscript{YOLO} with Corrector-Blender Model. Green blocks represent the parts being trained.}
	\label{fig:task1}
\end{figure}

\subsection{Method}
Task 1 is focused on training using a partially labeled dataset. For this purpose, we focused on employing a semi-supervised learning method to fully utilize the labeled and unlabeled parts of the dataset. We break our training process down into two parts: (1) supervised learning of MOD\textsubscript{YOLO}, and (2) semi-supervised training on the Corrector and Blender models.

As the first half of the training is done in a traditional fashion, we will only touch on the latter half in this section. An abstract figure of our method is shown in Figure \ref{fig:task1}. In our method, we introduced two new models: the Corrector Model and Blender Model. The semi-supervised training process is split into two parts: (1) unsupervised learning on the corrector model, and (2) supervised learning on the corrector and blender models. The two stages of training are done in the same order every training epoch, to reduce the risk of overfitting to either the labeled or unlabeled portions of the dataset.

In Stage 1 of the training process (shown in Figure \ref{fig:task1}a), the corrector model is being trained with the unlabeled parts of the dataset in an unsupervised manner. The loss being used in this stage is based on the constrained loss shown in the original ROAD-R paper \cite{DBLP:journals/ml/GiunchigliaSKCL23}. The loss is set as follows:
\begin{equation}
    \mathcal{L}_{S1} = \frac{1}{243}\sum^{243}_{i=1}t(r_i)
    \label{eq:cl}
\end{equation}
Here, $r_i$ corresponds to the $i$th requirement, and $t(r_i)$ corresponds to the fuzzy logic relaxations of $r_i$. Furthermore, we focused on using Product T-Norm for the fuzzy logic relaxation, and the predicted scores used in this loss are solely from the corrector model.

In Stage 2 of the training process (shown in Figure \ref{fig:task1}b), the corrector and blender models are being trained with the labeled parts of the dataset in a supervised manner. The loss is set as follows:
\begin{equation*}
    \mathcal{L}_{S2} = 10\times \mathcal{L}_{S1} + \mbox{BCELoss}(\hat{\boldsymbol{\mathit{y}}}_{C}, \boldsymbol{\mathit{y}}) + \mbox{BCELoss}(\hat{\boldsymbol{\mathit{y}}}_{B}, \boldsymbol{\mathit{y}})
\end{equation*}
Here, $\hat{\boldsymbol{\mathit{y}}}_{C}$ and $\hat{\boldsymbol{\mathit{y}}}_{B}$ correspond to the output predictions given by the corrector model and the blender model.

\subsection{Results}
\begin{table*}[t]
    \centering
    \caption{Comparison of models for Task 1}
    \begin{tabular}{@{\extracolsep{6pt}}cc}
        \hline\hline
        Model&Frame-mAP@0.5\\
        \hline
        Baseline Model&0.18\\
        MOD\textsubscript{YOLO}&0.2633\\
        MOD\textsubscript{YOLO} with Corrector-Blender Model&\textbf{0.2662}\\
        \hline\hline
    \end{tabular}
    \label{table:task1}
\end{table*}

The results for the comparison of models are shown in Table \ref{table:task1}. From the results, we can see that using the Corrector Model along with the Blender Model can have a positive impact on the overall performance. This shows that having minor changes to label confidence scores with respect to the requirements is important, even under normal object detection circumstances. Moreover, this demonstrates the positive impacts of using the unlabeled parts of the dataset to learn the requirements in an unsupervised manner.

\section{Task 2}
\subsection{Method}
Task 2 is focused on using the full dataset to train and finally outputting a set of labels that fully satisfy the requirements. To train the model that outputs confidence scores that directly lead to the correct output with the use of solvers, we use the constrained loss shown in Equation (\ref{eq:cl}). However, YOLO outputs confidence scores for each of its anchor points (total of $W\times H$ anchor points), making calculating the loss very expensive, both computation time-wise and resource-wise. Therefore, we calculate the constrained loss on anchor points that have at least one label with confidence scores above 0.5. Not only does this allow for less computation time and resources, but it also reduces the risk of trying to output high confidence scores for every bounding box, as the requirements do penalize empty label sets.

After the training of the model, we used MaxHS \cite{DBLP:conf/sat/HickeyB19}, a Partial Weighted MaxSAT solver, in an identical way to the original ROAD-R paper \cite{DBLP:journals/ml/GiunchigliaSKCL23}. This allowed us to output labels that are guaranteed to satisfy the given requirements. 

\subsection{Results}
\begin{table*}[t]
    \centering
    \caption{Comparison of models for Task 2}
    \begin{tabular}{@{\extracolsep{6pt}}cccc}
        \hline\hline
        Model&Precision@0.5&Recall@0.5&F1-Score@0.5\\
        \hline
        Baseline Model&0.49&0.34&0.40\\
        MOD\textsubscript{YOLO}&0.6769&0.4430&0.5355\\
        MOD\textsubscript{YOLO} with Constrained Loss &\textbf{0.7057}&\textbf{0.5405}&\textbf{0.6122}\\
        \hline\hline
    \end{tabular}
    \label{table:task2}
\end{table*}
The results for the comparison of models are shown in Table \ref{table:task2}. From the results, we can see that using constrained loss has a big effect on the overall performance of the model in terms of every metric. This shows that in models, learning to satisfy the requirements is needed if we were to use complete solvers to output sets of labels that are guaranteed to satisfy the requirements.

Another reason for the high scores is thought to be because of the agent-wise thresholding/NMS explained in Section 2. As at least one of the confidence scores of agent labels given to the MaxSAT solver is guaranteed to be above the threshold, it allows the solver to essentially not care about one of the requirements. Although the amount of leverage this method gives is not verified, we believe that there is a certain effect as the scores of MOD\textsubscript{YOLO} itself will place in 5th place of Task 2.

\section{Conclusion}
In this paper, we introduced MOD-CL, a framework that focuses on using constrained losses with multi-labeled object detection. We used MOD\textsubscript{YOLO}, a model based on the state-of-the-art object detection model YOLOv8, and implemented constrained losses in different forms to suit the separate tasks. In Task 1, we focused on modifying the outputs of YOLOv8 to give an output that satisfies the requirements more. From the results, we found that modifying the results in a particular way has a positive impact on the overall performance of the model. In Task 2, we focused on learning with constrained loss included to give outputs that lead to correct answers with the use of Weighted Partial MaxSAT solvers. The results show that the use of the constrained loss has a positive effect on the overall performance, suggesting its effectiveness in practical applications.

\begin{ack}
    This work has been supported by JSPS KAKENHI Grant Number JP21H04905 and JST CREST Grant Number JPMJCR22D3.
\end{ack}

\bibliographystyle{unsrt}
\bibliography{bibs}

\end{document}